\documentclass[11pt]{article}
\usepackage{coling2020}
\usepackage{times}
\usepackage{url}
\usepackage{latexsym}

\usepackage{graphicx}

\usepackage{multirow}

\colingfinalcopy % Uncomment this line for the final submission

\title{A Joint Model for Aspect-Category Sentiment Analysis with Shared Sentiment Prediction Layer}

\author{Yuncong Li \renewcommand{\thefootnote}{\fnsymbol{footnote}}\footnotemark[1], Zhe Yang \renewcommand{\thefootnote}{\fnsymbol{footnote}}\footnotemark[1], Cunxiang Yin, Xu Pan \renewcommand{\thefootnote}{\fnsymbol{footnote}}\footnotemark[2], \\
	\textbf{Lunan Cui,} \textbf{Qiang Huang} and \textbf{Ting Wei} \\
	Baidu Inc., Beijing, China \\
	{\tt\{liyuncong,yangzhe08,yincunxiang,panxu,cuilunan,huangqiang03,}\\
	{\tt weiting\}@baidu.com} \\
}

\date{}

\begin{document}
\maketitle
\begin{abstract}
   Aspect-category sentiment analysis (ACSA) aims to predict the aspect categories mentioned in texts and their corresponding sentiment polarities. Some joint models have been proposed to address this task. Given a text, these joint models detect all the aspect categories mentioned in the text and predict the sentiment polarities toward them at once. Although these joint models obtain promising performances, they train separate parameters for each aspect category and therefore suffer from data deficiency of some aspect categories. To solve this problem, we propose a novel joint model which contains a shared sentiment prediction layer. The shared sentiment prediction layer transfers sentiment knowledge between aspect categories and alleviates the problem caused by data deficiency. Experiments conducted on SemEval-2016 Datasets demonstrate the effectiveness of our model.
\end{abstract}

\renewcommand{\thefootnote}{\fnsymbol{footnote}}
\footnotetext[1]{Equal contribution.}
\footnotetext[2]{Corresponding author.}

\section{Introduction}
Aspect-category sentiment analysis (ACSA) is a subtask of aspect-based sentiment analysis (ABSA)~\cite{pontiki-etal-2014-semeval,pontiki2015semeval,pontiki2016semeval}. ACSA aims to identify all the aspect categories mentioned in texts and their corresponding sentiment polarities. An aspect category (or simply aspect) is an entity E and attribute A pair, denoted by E\#A. For example, in the text ``The place is small and cramped but the food is fantastic.'',  the aspect categories mentioned in the text are \emph{AMBIENCE\#GENERAL} and \emph{FOOD\#QUALITY}, and their sentiment polarities are negative and negative, respectively. 

Many methods have been proposed to address the ACSA task. However, most existing methods~\cite{zhou2015representation,movahedi2019aspect,wang2016attention,ruder2016hierarchical,cheng2017aspect,xue2018aspect,tay2018learning} divide the ACSA task into two subtasks: aspect category detection (ACD) which detects aspect categories in a text and sentiment classification (SC) which categorizes the sentiment polarities with respect to the detected aspect categories, and perform these two tasks separately. Such two-stage approaches lead to error propagation, that is, errors caused by aspect category detection would affect sentiment classification. To avoid error propagation, previous studies~\cite{schmitt2018joint,hu2019can,wang2019aspect} have proposed some joint models, which jointly model the detection of aspect categories and the classification of their polarities. Further more, given a text, these joint models detect all the aspect categories mentioned in the text and predict the sentiment polarities toward them at once. Although these joint models obtain promising performances, they train separate parameters for each aspect category and therefore suffer from data deficiency of some aspect categories. For example, the english laptops domain dataset from SemEval-2016 task 5: Aspect-based Sentiment Analysis~\cite{pontiki2016semeval} has a quarter of the aspect categories whose sample size are less than or equal to 2 (see Table~\ref{Statistics}).  Previous joint models will under-fit on the aspect categories with deficient samples.

To solve the problem caused by data deficiency mentioned above, we propose a novel joint model, which contains a shared sentiment prediction layer. Our model is based on the observation that the sentiment expressions and their polarities of different aspect categories are transferable. For instance, in Table~\ref{example-for-shared-sentiment-prediction-layer}, the three aspect categories \emph{LAPTOP\#QUALITY}, \emph{LAPTOP\#PRICE}, and \emph{LAPTOP\#OPERATION\_PERFORMANCE} have the same sentiment word “surprised” and the consistent polarity. The shared sentiment prediction layer transfers sentiment knowledge between aspect categories and alleviates the problem caused by data deficiency.

In summary, the main contributions of our work are as follows:
\begin{itemize}
	\item We propose a novel joint model for the aspect category sentiment analysis (ACSA) task, which contains a shared sentiment prediction layer. The shared sentiment prediction layer transfers sentiment knowledge between aspect categories and alleviates the problem caused by data deficiency.
	\item Experiments conducted on SemEval-2016 Datasets demonstrate the effectiveness of our model.
\end{itemize}

\begin{table}
	\begin{center}
		\begin{tabular}{|l|l|l|}
			\hline
			\textbf{Aspect Category} & \textbf{Text} & \textbf{Polarity} \\
			\hline
			LAPTOP\#QUALITY & \multirow{2}{16em}{...I was \textbf{surprised}  at  the overall quality and the price...} & positive \\
			\cline{1-1}\cline{3-3}
			LAPTOP\#PRICE &  & positive \\
			\hline
			LAPTOP\#OPERATION\_PERFORMANCE & \multirow{2}{16em}{...I was \textbf{surprised} with the performance and quality of this HP Laptop...} & positive \\
			\cline{1-1}\cline{3-3}
			LAPTOP\#QUALITY &  & positive \\
			\hline
		\end{tabular}
	\end{center}
	\caption{\label{example-for-shared-sentiment-prediction-layer}Different aspect categories have the same sentiment word which have the same sentiment polarity.}
\end{table}

\section{Related Work}
Existing methods for Aspect-Category Sentiment Analysis (ACSA) can be divided into two categories: two-stage methods and joint models.

\textbf{Two-stage methods} perform the ACD task and the SC task separately. Zhou et al.~\shortcite{zhou2015representation} and Movahedi et al.~\shortcite{movahedi2019aspect} perform the ACD task. Zhou et al.~\shortcite{zhou2015representation} propose a semi-supervised word embedding algorithm to obtain word embeddings on a large set of reviews, which are then used to generate deeper and hybrid features to predict the aspect category. Movahedi et al.~\shortcite{movahedi2019aspect} utilize topic attention to attend to different aspects of a given text. Many methods~\cite{wang2016attention,ruder2016hierarchical,cheng2017aspect,xue2018aspect,tay2018learning,liang2019novel,jiang2019challenge,xing2019earlier,sun2019utilizing,zhu2019aspect,lei2019human} have been proposed for the SC task. Wang et al.~\shortcite{wang2016attention} first propose aspect embedding (AE) and use an Attention-based Long Short-Term Memory Network (AT-LSTM) to generate aspect-specific text representations for sentiment classification based on aspect embedding. Ruder et al.~\shortcite{ruder2016hierarchical} propose a hierarchical bidirectional LSTM (H-LSTM) to modeling the interdependencies of sentences in a review. Tay et al.~\shortcite{tay2018learning} propose a method named Aspect Fusion LSTM (AF-LSTM) to model word-aspect relationships. Xue and Li~\shortcite{xue2018aspect} propose a model, namely Gated Convolutional network with Aspect Embedding (GCAE), which incorporates aspect information into the neural model by gating mechanisms. Jiang et al.~\shortcite{jiang2019challenge} proposed new capsule networks to model the complicated relationship between aspects and contexts. All the two-stage methods have the problem of error propagation.

\textbf{Joint models} jointly model the detection of aspect categories and the classification of their polarities. Only a few joint models~\cite{schmitt2018joint,hu2019can,wang2019aspect} have been proposed for ACSA. Schmitt et al.~\cite{schmitt2018joint} propose two joint models: End-to-end LSTM and End-to-end CNN, which produce all the aspect categories and their corresponding sentiment polarities at once. Hu et al.~\shortcite{hu2019can} propose constrained attention networks (CAN), which extends AT-LSTM to multi-task settings and introduces orthogonal and sparse regularizations to constrain the attention weight allocation. As a result, the CAN achieves better sentiment classification performance. However, to train the CAN, we need to annotate the multi-aspect sentences with overlapping or nonoverlapping. Wang et al.~\shortcite{wang2019aspect} propose the aspect-level sentiment capsules model (AS-Capsules), which utilizes the correlation between aspect category and sentiment through shared components including capsule embedding, shared encoders, shared attentions and a shared recurrent neural network. These joint models train separate parameters for each aspect category, which results in that these models under-fit on the aspect categories with deficient samples.

\section{Proposed Model}
\begin{figure}
	\centering
	\includegraphics[width=0.9\textwidth]{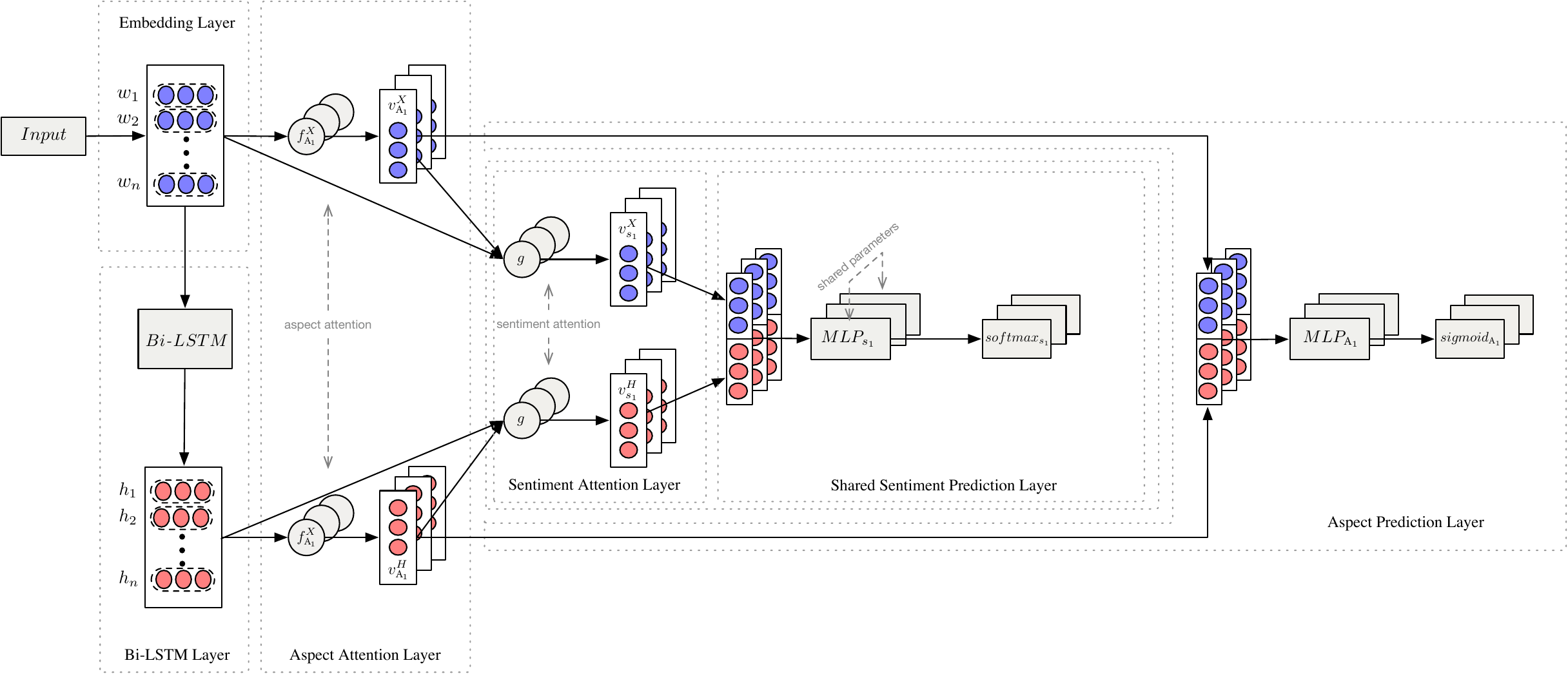}
	\caption{Overall architecture of the proposed method.} \label{model}
\end{figure}

We first formulate the problem. There are $N$ predefined aspect categories $A= \left\{A_1,A_2,\cdots, A_N \right\}$ and $M$ predefined sentiment polarities $P=\left\{P_1,P_2,\cdots,P_M \right\}$ in the dataset. Given a sentence or a review, denoted by $S=\left\{w_1,w_2,\cdots, w_n \right\}$, the task aims to predict the aspect categories and the 
corresponding sentiment polarities, i.e., aspect-sentiment pairs$ \left\{<A_j,P_k>\right\}$, expressed in the text. 
The overall model architecture is illustrated in Figure~\ref{model}, which contains six modules: embedding layer, Bi-LSTM layer, aspect attention layer, sentiment attention layer, aspect category prediction layer, and shared sentiment prediction layer. Then, we display the details of each module and introduce the training objective function.

\subsection{Embedding Layer}
The input to our model is a text consisting of $n$ words  $\left\{w_1,w_2,...,w_n \right\} $. With a embedding matrix  $U$, the input text is converted to a sequence of vectors  $X=\left\{x_1,x_2,...,x_n\right\} $. Where $U\in R^{d_w\times|V| } $ , $d_w$ is the dimension of the word embeddings, and $|V|$is the vocabulary size. 

\subsection{Bidirectional LSTM Layer}
The word embeddings of the text are then fed into a Bidirectional LSTM~\cite{graves2013speech} network (Bi-LSTM) with two LSTM~\cite{hochreiter1997long} networks. We can obtain two hidden representations, and then concatenate the forward hidden state and backward hidden state of each word. Formally, given the sequence of vectors  $X=\left\{x_1,x_2,\cdots,x_n\right\}$, Bi-LSTM outputs hidden states $H=\left\{h_1,h_2,\cdots,h_n\right\}$. At each time step $i=1,2,\cdots,n$, the hidden state $h_i$  of the Bi-LSTM is computed by:

\begin{equation}
	\overrightarrow{h_i} =  \overrightarrow{LSTM} (\overrightarrow{h}_{i-1}, x_i)
\end{equation}
\begin{equation}
	\overleftarrow{h_i} = \overleftarrow{LSTM}( \overleftarrow{h}_{i+1}, x_i )
\end{equation}
\begin{equation}
	h_i =[ \overrightarrow{h_i},\overleftarrow{h_i} ]
\end{equation}
where $\overrightarrow{h _i}\in R^{d_s }$,$\overleftarrow{h _i}\in R^{d_s} $, $h_i\in R^{2d_s }$, and $d_s$ denotes the size of the hidden state of LSTM.

\subsection{Aspect Attention Layer}
This layer applies an attention mechanism on the outputs of both the embedding layer and the Bi-LSTM layer and generates aspect-specific representations for the ACD task. Different aspect categories have different attention parameters. The process can be formulated as follows:
\begin{equation}
	v^X_{A_j}= f^X _{A_j}(X),j=1,\cdots, N
\end{equation}
\begin{equation}
	v^H_{A_j}= f^H _{A_j}(H),j=1,\cdots, N
\end{equation}
where f($\cdot$) is an attention mechanism~\cite{yang2016hierarchical} and can be defined as follows:
\begin{equation}
	f(V) = v = \Sigma^n_{i=1}\alpha_i v_i 
\end{equation}
\begin{equation}
	u_i = tanh(W_a v_i+b_a )\quad for\quad i=1,2,\cdots,n
\end{equation}
\begin{equation}
	\alpha_i =\frac{  exp(u^T_iu_w)}{(\Sigma^n_{j=1}exp(u^T_j u_w)  }\quad for \quad i=1,2,\cdots,n 
\end{equation}
where $V=\left\{v_1,\cdots,v_i,\cdots,v_n \right\}$ is a sequence of vectors and $v_i\in R^d$. $W_a \in R^{m\times d}$,$b_a\in R^m$, and $u_w\in R^m$ are the parameters of the attention mechanism. $m$ is the dimensionality of the attention context vector, and $d$ is the dimensionality of the input vector. Note that the vector $v$ generated by $f(V)$ is a weighted sum of vectors in $V$ and is in the same semantic space with them.

\subsection{Aspect Category Prediction Layer}
Aspect category prediction layer takes as input the concatenation of the aspect-specific representations at the embedding layer and the Bi-LSTM layer for the ACD task and predicts whether the text mentions the aspect categories. Formally, for the $j$-th aspect category:
\begin{equation}
	v_{A_j}=[v_{A_j}^X, v^H_{A_j}]
\end{equation}
\begin{equation}
	\widehat{y} _{A_j}=\sigma(\widehat{W} _{A_j} ReLU(W_{A_j} v_{A_j} +b_{A_j}  )+\widehat{b}_{A_j})
\end{equation}
\begin{equation}
	\sigma(x)=  \frac{1}{(1+e^{-x}) }
\end{equation}
where $W_{A_j}$,$ b_{A_j}$, $\widehat{W} _{A_j}$, and $\widehat{b}_{A_j}$ are the parameters of the $j$-th aspect category. If $\widehat{y} _{A_j}$ is greater than the specified  threshold $\tau$, we judge that the $j$-th aspect category is mentioned by the text.  

\subsection{Sentiment Attention Layer}
This layer generates aspect-specific text representations for the SC task based on aspect-specific text representations for the ACD task. For the $j$-th aspect category, its aspect-specific text representations for the SC task can be computed as follows:
\begin{equation}
	v_{s_j}^X=g(X,v_{A_j}^X  ) 
\end{equation}
\begin{equation}
	v_{s_j}^H=g(H,v_{A_j}^H  )
\end{equation}
where $X$ and $H$ are the outputs of the embedding layer and the Bi-LSTM layer, respectively. $ v^X_{A_j}$and $v^H_{A_j}$ are the aspect-specific text representations of the $j$-th aspect category for the ACD task at the outputs of the embedding layer and the Bi-LSTM layer respectively. g($\cdot$) is an attention mechanism~\cite{vaswani2017attention} and can be defined as follows:
\begin{equation}
	v_s=g(X,v_q )
\end{equation}
\begin{equation}
	\beta_i =\frac{  exp(x^T_iv_q)}{\Sigma^n_{j=1}x^T_j v_q }\quad for \quad i=1,2,\cdots,n 
\end{equation}
\begin{equation}
	v_s=\Sigma^n_{i=1}\beta_i x_i
\end{equation}
where $X$ is a sequence vectors $\left\{x_1,x_2,\cdots,x_n\right\}$, and $v_q$ is the query vector of the attention. We use the dot product to compute attention weights because it does not import extra aspect-specific parameters. Since the query vector and the key vector of the attention are in the same semantic space in our model, the dot product is reasonable.

\subsection{Shared Sentiment Prediction Layer}
The aspect-specific text representation of the $j$-th aspect for the SC task is generated by concatenating the aspect-specific text representations of the $j$-th aspect category for the SC task at the outputs of the embedding layer and the Bi-LSTM layer. The representation are then fed to a fully connected layer with the ReLU activation function and then the output of the fully connected layer is fed to another fully connected layer with the softmax activation function to generate sentiment probability distribution. Formally, for the $j$-th aspect category:
\begin{equation}
	v_{s_j} =[v_{s_j}^X,v_{s_j}^H ]
\end{equation}
\begin{equation}
	\widehat{y}_{s_j} = softmax(\widehat{W}_s ReLU(W_s v_{s_j} +b_s )+\widehat{b}_s)
\end{equation}
\begin{equation}
	softmax(x)_i = \frac{exp(x_i)}{\Sigma^M_{k=1}exp(x_k)}
\end{equation}
where $W_s$ , $b_s$ , $\widehat{W}_s$ , and $\widehat{b}_s$ are the shared parameters of all aspect categories.

\subsection{ Loss}
For the aspect category detection task, as each prediction is a binary classification problem, the loss function is defined by:
\begin{equation}
	L_A (\theta)=-\Sigma^N_{j=1}y_{A_j }  log\widehat{y} _{A_j } +(1-y_{A_j })log(1-\widehat{y}_{A_j} ) 
\end{equation} 
For the sentiment classification task, the loss function is defined by:
\begin{equation}
	L_s (\theta) =-\Sigma^N_{j=1}\Sigma^M_{k=1}y_{s_{j_k}}  log(\widehat{y} _{s_{j_k}})
\end{equation} 
if the $j$-th aspect category is not mentioned in the text, $y_{s_{j_k}} =0$ for $k=1,2,\cdots,M$.

We jointly train our model for the two tasks. The parameters in our model are then trained by minimizing the combined loss function:
\begin{equation}
	L_s (\theta)=L_A (\theta) + \eta L_s (\theta) +\lambda ||\theta||^2_2
\end{equation} 
where $\eta$ is the weight of sentiment classification loss, $\lambda$  is the L2 regularization factor and $\theta$ contains all the parameters except for Bi-LSTM layer's parameters. Furthermore, to avoid over-fitting, we adopt the dropout strategy to enhance our model.

\section{Experiments}
\subsection{Datasets}
We conduct experiments on four public datasets from SemEval-2016 task 5: Aspect-based Sentiment Analysis~\cite{pontiki2016semeval}:

\textbf{CH-CAME-SB1} is a Chinese sentence-level dataset about digital cameras domain.

\textbf{CH-PHNS-SB1} is a Chinese sentence-level dataset about mobile phones domain.

\textbf{EN-REST-SB2} is an English review-level dataset about restaurants domain.

\textbf{EN-LAPT-SB2} is an English review-level dataset about laptops domain.

We randomly split the original training set into training, validation sets in the ratio 9:1. We use quartiles to measure the distribution of the sample size of aspects in these datasets. Detailed statistics are summarized in Table~\ref{Statistics}. Particularly, for the three datasets, CH-CAME-SB1, CH-PHNS-SB1, and EN-LAPT-SB2, the sample size of 50\% of the aspects are no more than 7.

\begin{table}
	\begin{center}
		\begin{tabular}{|l|l|l|l|l|l|l|l|l|l|l|}
			\hline
			\textbf{Dataset} & \textbf{\#aspect} & \textbf{\#polarity} & \textbf{\#train} & \textbf{\#val} & \textbf{\#test} & \textbf{\#min} & \textbf{\#Q1} & \textbf{\#Q2} & \textbf{\#Q3} \\
			\hline
			CH-CAME-SB1 & 75 & 2 & 1090 & 169 & 481 & 1.0 & 1.0 & 6.0 & 13.0\\
			\hline
			CH-PHNS-SB1 & 81 & 2 & 1152 & 181 & 529 & 1.0 & 1.8 & 4.5 & 23.3\\
			\hline
			EN-LAPT-SB2 & 88 & 4 & 355 & 40 & 80 & 1.0 & 2.0 & 7.0 & 20.0\\
			\hline
			EN-REST-SB2 & 12 & 4 & 301 & 34 & 90 & 20.0 & 36.5 & 68.5 & 177.0\\
			\hline
		\end{tabular}
	\end{center}
	\caption{\label{Statistics}  Statistics of the datasets. \#aspect and \#polarity represent the number of predefined aspects and sentiment polarities, respectively. \#train, \#dev, and \#test represent the sample size of training sets, validation sets, and test sets, respectively. \#min indicates the minimum value of the aspect sample size. \#Q1, \#Q2, \#Q3 are the first quartile, the second quartile, and the third quartile of the aspect sample size, respectively.}
\end{table}

\subsection{Evaluation Metrics}
We use micro-averaged F1-scores as the evaluation metric for both the ACSA and the ACD:
\begin{equation}
	F_1=\frac{2*P*R}{P+R}
\end{equation}
where precision and recall are defined as:
\begin{equation}
	P=\frac{|S \cap G|}{|S|}
\end{equation}
\begin{equation}
	R=\frac{|S \cap G|}{|G|}
\end{equation}
Here $S$ is the set of aspect-sentiment pairs or aspect category annotations (in ACSA and ACD, respectively) that a model returns for all the test texts, and $G$ is the set of the gold (correct) aspect-sentiment pairs or aspect category annotations. 
To evaluate the SC task, we use the gold aspect category annotations to select sentiment polarities model predicts and calculated the accuracy.

\subsection{Comparison Methods}
We select the following methods for comparison.

\textbf{End-to-end LSTM}~\cite{schmitt2018joint} performs the ACSA task, which jointly models the detection of aspects and the classification of their polarities in an end-to-end trainable neural network.

\textbf{End-to-end CNN}~\cite{schmitt2018joint} is an CNN version of End-to-end LSTM, which replaces the Bi-LSTM in End-to-end LSTM with a convolutional neural network (CNN) described in~\cite{kim-2014-convolutional}. 

\textbf{AS-Capsules}~\cite{wang2019aspect} utilizes the correlation between aspect category and sentiment through shared components including capsule embedding, shared encoders, shared attentions and a shared recurrent neural network.

\textbf{SemEval-2016 Best} is the best model for each subtask of SemEval-2016 task 5: Aspect based Sentiment Analysis~\cite{pontiki2016semeval}.

\textbf{Our Model – w/o Share} was added to show the effectiveness of the shared sentiment prediction layer, which trains a separate sentiment prediction layer for each aspect.

\subsection{Implementation Details}
We implement all models in Keras. We set $\lambda=0.01$ and gradient clipping norm to 5. Adam~\cite{kingma2014adam} optimizer is applied to minimize the loss. We apply a dropout of $p=0.5$ after the embedding layer and the Bi-LSTM layer. Hidden layer size for Bi-LSTM is 100. We use 300-dimensional word embeddings. We use GloVe~\cite{pennington2014glove} embeddings which are pre-trained on an unlabeled corpus whose size is about 840 billion for English and Skip-Gram~\cite{mikolov2013distributed} embeddings which are pre-trained on the Baidu Encyclopedia dataset for Chinese. If an aspect is not mentioned, its corresponding sentiment label is set to a zero vector. We set threshold $\tau=0.25$ for aspect category detection. While batch size is 32 on CAME-SB1 and CH-PHNS-SB1, batch size is 10 on EN-LAPT-SB2 and EN-REST-SB2. The sentiment classification loss weight is 1 on CH-CAME-SB1, CH-PHNS-SB1, and EN-LAPT-SB2, and is 0.6 on EN-REST-SB2. To reduce the randomness of results, we train each model three times and report their averaged scores. 
\begin{table}
	\begin{center}
		\begin{tabular}{|l|l|l|l|l|}
			\hline
			\textbf{Models} & \textbf{CH-CAME-SB1} & \textbf{CH-PHNS-SB1} & \textbf{EN-LAPT-SB2} & \textbf{EN-REST-SB2} \\
			\hline
			End-to-end cnn & 34.96 & 19.87 & 36.52 & 66.03 \\
			\hline
			End-to-end lstm & 41.52 & 26.30 & 37.94 & 63.75 \\
			\hline
			AS-Capsules & 38.85 & 27.05 & 33.47 & 63.99 \\
			\hline
			Our Model & \textbf{42.01} & \textbf{28.98} & \textbf{50.05} & 68.24 \\
			– w/o  Share & 36.23 & 22.72 & 49.43 & \textbf{68.28} \\
			\hline
		\end{tabular}
	\end{center}
	\caption{\label{results-of-ACSA} Results of the ACSA task in terms of micro-averaged F1-scores(\%).}
\end{table}
\begin{table}
	\begin{center}
		\begin{tabular}{|l|l|l|l|l|}
			\hline
			\textbf{Models} & \textbf{CH-CAME-SB1} & \textbf{CH-PHNS-SB1} & \textbf{EN-LAPT-SB2} & \textbf{EN-REST-SB2} \\
			\hline
			SemEval-2016 Best & 80.45 & 73.34 & 75.05 & 81.93 \\
			\hline
			End-to-end cnn & 70.55 & 64.15 & 69.42 & 80.78 \\
			\hline
			End-to-end lstm & 75.12 & 67.36 & 72.00 & 80.03 \\
			\hline
			AS-Capsules & 76.96 & 71.56 & 69.05 & 76.794 \\
			\hline
			Our Model & \textbf{82.54} & \textbf{76.50} & \textbf{75.91} & 82.43 \\
			– w/o  Share & 69.09 & 59.86 & 70.53 & \textbf{83.33} \\
			\hline
		\end{tabular}
	\end{center}
	\caption{\label{results-of-SC}  Results of the SC task in terms of accuracy(\%).}
\end{table}
\begin{table}
	\begin{center}
		\begin{tabular}{|l|l|l|l|l|}
			\hline
			\textbf{Models} & \textbf{CH-CAME-SB1} & \textbf{CH-PHNS-SB1} & \textbf{EN-LAPT-SB2} & \textbf{EN-REST-SB2} \\
			\hline
			SemEval-2016 Best & 36.3 & 22.5 & 60.4 & \textbf{83.9} \\
			\hline
			End-to-end cnn & 47.83 & 26.64 & 42.25 & 76.20 \\
			\hline
			End-to-end lstm & \textbf{52.98} & 33.81 & 43.81 & 76.24 \\
			\hline
			AS-Capsules & 48.72 & 33.66 & 40.99 & 78.29 \\
			\hline
			Our Model & 51.81 & \textbf{36.67} & \textbf{62.91} & 81.65 \\
			– w/o  Share & 52.16 & 36.54 & 62.72 & 81.45 \\
			\hline
		\end{tabular}
	\end{center}
	\caption{\label{results-of-acd} Results of the ACD task in terms of micro-averaged F1-scores(\%).}
\end{table}

\subsection{Results}
Table~\ref{results-of-ACSA}, Table~\ref{results-of-SC}, and Table~\ref{results-of-acd} show our experimental results on the ACSA, SC, and ACD tasks, respectively. The best results are marked in bold. 

Table~\ref{results-of-ACSA} shows the experimental results on the ACSA task, which show the overall performance of our joint model. We observe that our proposed joint model outperforms the baseline models on all datasets, which demonstrates the effectiveness of our model.

The experimental results on the SC task are in Table~\ref{results-of-SC}. First, we observe that our model surpasses all baseline models on all datasets, which indicates the effectiveness of our model predicting the sentiment polarities toward given aspect categories. Second, our model outperforms its variant (– w/o  Share) by 13.45\%, 16.64\% and 5.38\% on CH-CAME-SB1, CH-PHNS-SB1, and EN-LAPT-SB2 datasets, respectively. The reason is that the three datasets have many aspect categories which only have a few instances and benefit from parameter sharing. This shows that our shared parameter prediction layer can alleviate the problem caused by data deficiency. Meanwhile, our model obtains worse perfromance than its variant (– w/o  Share) on the EN-REST-SB2 dataset. The possible reason is that the sample size of the aspect categories in the EN-REST-SB2 dataset is enough to train independent sentiment prediction parameters, and parameter sharing brings some noise between aspect categories. 

Table~\ref{results-of-acd} shows the results on the ACD task. Although we did not specifically optimize our model for the ACD task, our model still achieves competitive performance. Specifically, our model outperforms all baselines on the CH-PHNS-SB1 and EN-LAPT-SB2 datasets.

\subsection{Case Studies}
To have an intuitive understanding of our proposed shared sentiment prediction layer for the SC task, we use the EN-LAPT-SB2 dataset to illustrate the impact of knowledge transferring. The selected aspect category from the dataset is \emph{OS\#MISCELLANEOUS}. There are only two samples with both negative polarities in the training set, while there are two samples in the test set, whose polarities are negative and positive, respectively.  Table~\ref{case} shows that our model can correctly predict the polarity of the sample with positive sentiment in the test set. After removing the shared sentiment prediction layer, – w/o  Share fails to predict the polarity of the sample with positive sentiment, which confirms the importance of the shared sentiment prediction layer.

\begin{table}
	\begin{center}
		\begin{tabular}{|l|l|l|l|l|}
			\hline
			& \multirow{2}*{\textbf{Text}} & \multirow{2}*{\textbf{Label}} & \multirow{2}{3em}{\textbf{Our model}} & \multirow{2}*{\textbf{-w/o Share}} \\
			&  &  &  & \\
			\hline
			\multirow{5}{4em}{Train set} & \multirow{3}{18em}{...The only objection I have is that after you buy it the windows 7 system is a starter and charges for the upgrade...} & \multirow{3}*{negtive} & & \\
			&  &  &  & \\
			&  &  &  & \\
			\cline{2-5}
			& \multirow{2}{18em}{...The flaws are, this computer is not for computer gamers because of the OS X...} & \multirow{2}*{negtive} & & \\
			&  &  &  & \\
			\hline
			\multirow{2}{4em}{Test set} & ...The OS is easy, and offers all kinds of surprises... & positive & positive & negtive \\
			\cline{2-5}
			& ...The free upgrade to Mountain Lion FAILED... & negtive & negtive & negtive \\
			\hline
		\end{tabular}
	\end{center}
	\caption{\label{case}Impact of the shared sentiment prediction layer on the sentiment prediction of the aspect category \emph{OS\#MISCELLANEOUS}.}
\end{table}

\section{Conclusion}
In this work, we propose a novel joint model which contains a shared sentiment prediction layer. The shared sentiment prediction layer transfers sentiment knowledge between aspect categories and alleviates the problem caused by data deficiency. Experiments conducted on four datasets from SemEval-2016 task 5 demonstrate the effectiveness of our model. Furture work could consider introducing extra component that prevents the shared sentiment prediction layer from transfering aspect-specific sentiment knowledge.

\bibliographystyle{coling}
\bibliography{ccl2020-en}

\end{document}